\documentclass{article}

\usepackage{microtype}
\usepackage{graphicx}
\usepackage{subcaption}
\usepackage{booktabs} 
\usepackage{cuted} 
\usepackage{caption} 
\usepackage{xcolor}

\usepackage{hyperref}

\usepackage[preprint]{icml2026}

\usepackage{amsmath}
\usepackage{amssymb}
\usepackage{mathtools}
\usepackage{amsthm}
\usepackage[table]{xcolor}
\usepackage{booktabs}
\usepackage{colortbl}

\definecolor{osfirst}{RGB}{110,170,230}   
\definecolor{ossecond}{RGB}{165,205,245}  
\definecolor{osthird}{RGB}{220,235,250}   

\definecolor{csfirst}{RGB}{247,176,103}   
\definecolor{cssecond}{RGB}{250,210,122}  
\definecolor{csthird}{RGB}{252,237,170}   

\usepackage[capitalize,noabbrev]{cleveref}

\theoremstyle{plain}

\theoremstyle{definition}

\theoremstyle{remark}

\usepackage[textsize=tiny]{todonotes}

\usepackage{xcolor}  
\usepackage{hyperref} 

\hypersetup{
    colorlinks=true,     
    linkcolor=red,       
    urlcolor={rgb,255:red,255;green,50;blue,150}         
}

\begin{document}

\twocolumn[
  \icmltitle{AnimationBench: Are Video Models Good at Character-Centric Animation?}



  \icmlsetsymbol{equal}{*}
  \icmlsetsymbol{corresponding}{$\dagger$}
  \icmlsetsymbol{corr_lead}{$\dagger\ddagger$}

  \begin{icmlauthorlist}
    \icmlauthor{Leyi Wu}{HKUST,HKUSTGZ,equal}
    \icmlauthor{Pengjun Fang}{HKUST,equal}
    \icmlauthor{Kai Sun}{HKUST,equal}
    \icmlauthor{Yazhou Xing}{HKUST,NewAILabs,corr_lead}
        
    \icmlauthor{Yinwei Wu}{HKUST}
    \icmlauthor{Songsong Wang}{HKUST}
    \icmlauthor{Ziqi Huang}{NTU}
    \icmlauthor{Dan Zhou}{PearlStudio}
    \icmlauthor{Yingqing He}{HKUST,NewAILabs}
    \icmlauthor{Ying-Cong Chen}{HKUST,HKUSTGZ}
    \icmlauthor{Qifeng Chen}{HKUST,NewAILabs,corresponding}
  \end{icmlauthorlist}

  \icmlaffiliation{HKUST}{HKUST}
  \icmlaffiliation{HKUSTGZ}{HKUST(GZ)}
  \icmlaffiliation{NTU}{NTU}
  \icmlaffiliation{NewAILabs}{New AI Labs}
   \icmlaffiliation{PearlStudio}{Pearl Studio}

  \icmlcorrespondingauthor{Yazhou Xing}{yzxing87@gmail.com}
  \icmlcorrespondingauthor{Qifeng Chen}{cqf@ust.hk}

  \icmlkeywords{Video Generation, Cartoon Content Creation, Benchmark, Visual Language Models}

  \vskip 0.3in
]



\printAffiliationsAndNotice{\icmlEqualContribution, $\ddagger$ Project Lead, $\dagger$ Corresponding Author}  

\begin{abstract}
Video generation has advanced rapidly, with recent methods producing increasingly convincing animated results. 
However, existing benchmarks—largely designed for realistic videos—struggle to evaluate animation-style generation with its stylized appearance, exaggerated motion, and character-centric consistency.
Moreover, they also rely on fixed prompt sets and rigid pipelines, offering limited flexibility for open-domain content and custom evaluation needs.
To address this gap, we introduce AnimationBench, the first systematic benchmark for evaluating animation image-to-video generation. AnimationBench operationalizes the \textit{Twelve Basic Principles of Animation} and \textit{IP Preservation} into measurable evaluation dimensions, together with \textit{Broader Quality Dimensions} including semantic consistency, motion rationality, and camera motion consistency.
The benchmark supports both a standardized close-set evaluation for reproducible comparison and a flexible open-set evaluation for diagnostic analysis, and leverages visual-language models for scalable assessment. 
Extensive experiments show that AnimationBench aligns well with human judgment and exposes animation-specific quality differences overlooked by realism-oriented benchmarks, leading to more informative and discriminative evaluation of state-of-the-art I2V models.
The project webpage is
~\textcolor{pink}{\href{https://animationbench.github.io/}{\texttt{https://animationbench.github.io/}}}.

\end{abstract}

\begin{figure}[t!]
    \centering
    \includegraphics[width=\linewidth]{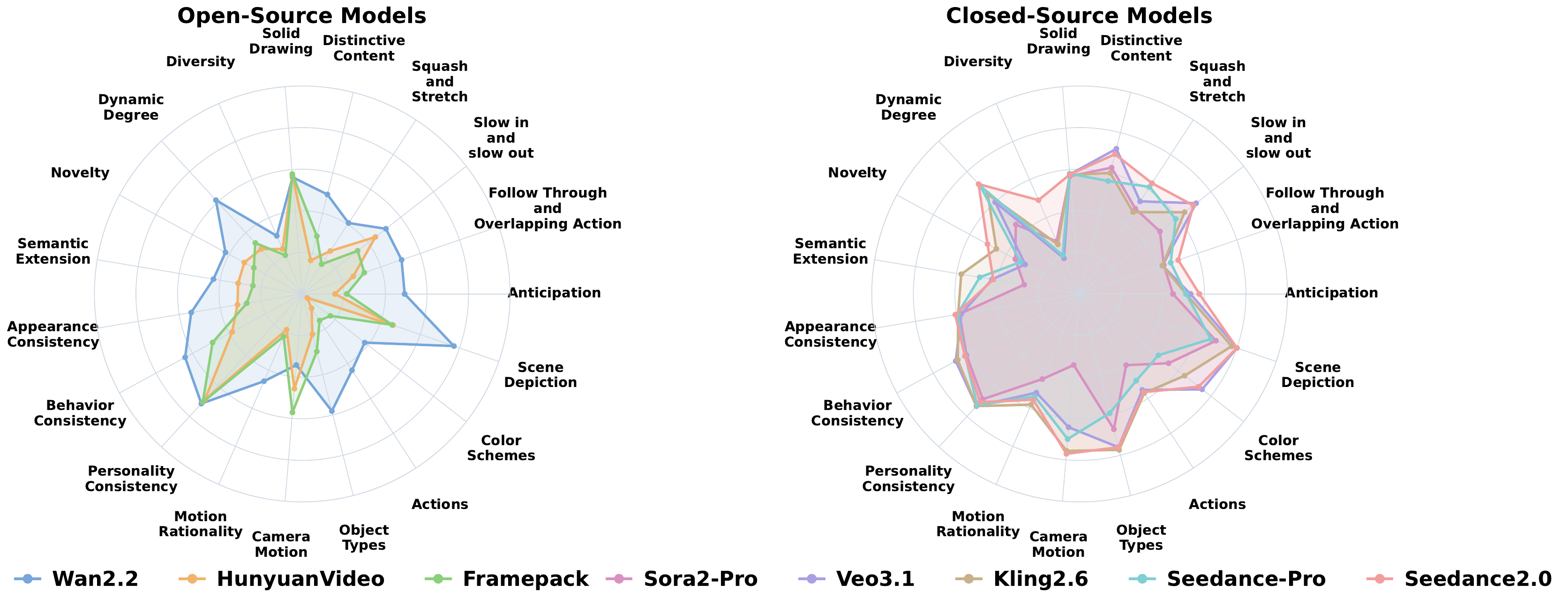}
    \caption{\textbf{AnimationBench Evaluation Results.}
    We visualize the evaluation results of seven video generation models (including both open-source and closed-source models) across all 19 AnimationBench dimensions.
    For better visualization, we normalize scores per dimension.
    For comprehensive numerical results, please refer to Table~\ref{tab:overall}.}
    \label{fig:overall score}
    \vspace{-8mm}
\end{figure}

\section{Introduction}

\begin{figure*}
    \centering
  \includegraphics[width=0.95\textwidth]{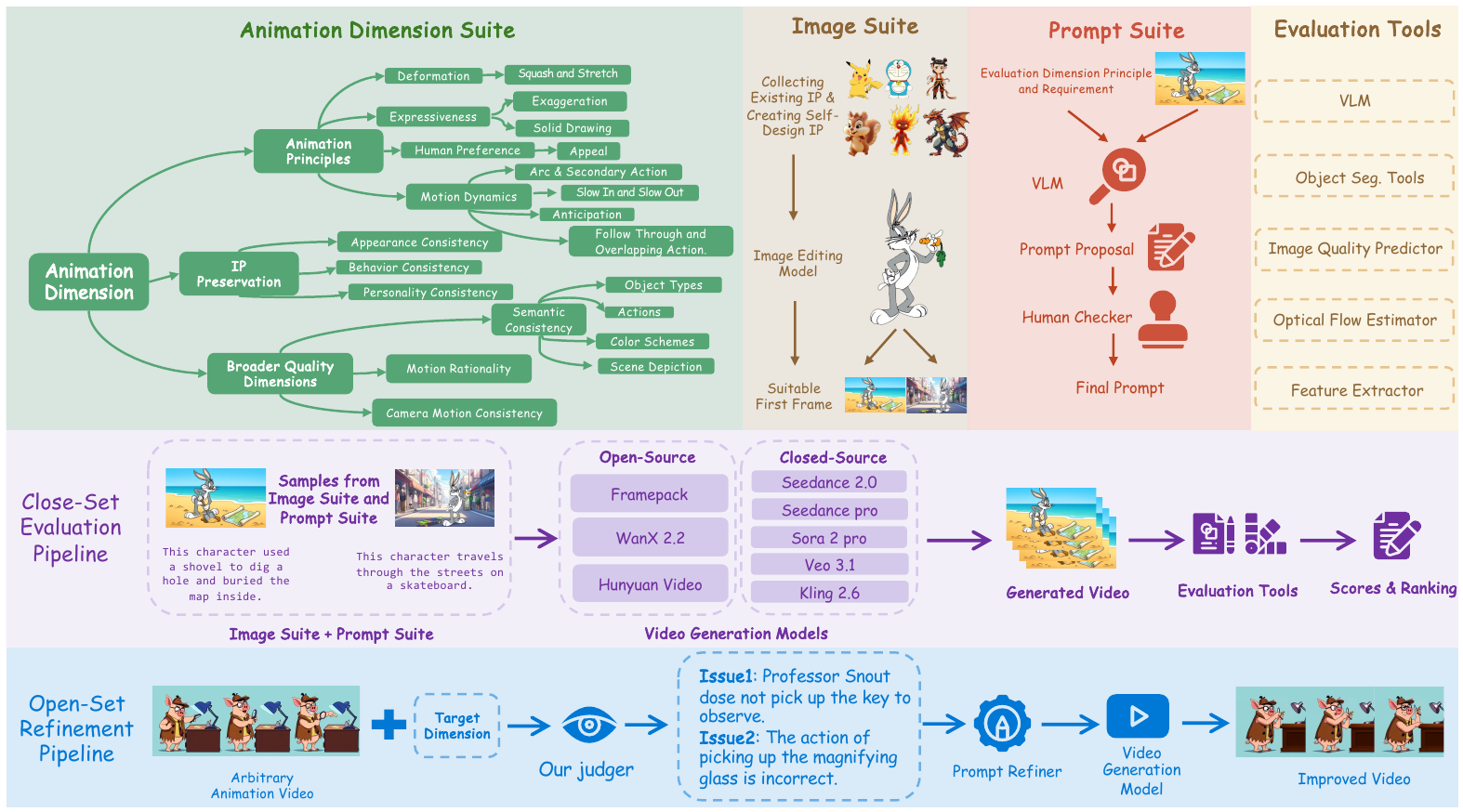}
  \captionof{figure}{\textbf{Overview of AnimationBench.}
  AnimationBench provides a principled benchmark suite for animation video generation.
  We organize evaluation into a hierarchical \textbf{Animation Dimension Suite} (IP Preservation, Animation Principles, and Broader Quality Dimensions), and construct paired \textbf{Image Suite} and \textbf{Prompt Suite} to drive IP-conditioned generation.
  Most dimensions are assessed with a unified, VLM-based \textbf{Evaluation Tool Suite}, complemented by specialized evaluators when needed.
  We support two usage modes: a standardized \textbf{close-set} pipeline for reproducible comparison across models, and an \textbf{open-set} diagnostic/refinement pipeline for evaluating arbitrary animation videos and guiding iterative prompt refinement.}
  \label{fig:overview}
  \vspace{-6mm}
\end{figure*}

Recent advances in video generation models~\cite{wu2025hunyuanvideo, kuaishou2024klingai, openai2024sora, chen2025seedance, veo2025, zhang2025packing, wan2025wan} have enabled the synthesis of increasingly high-quality and temporally coherent videos, marking a major step toward general-purpose visual content creation.
Beyond photorealistic videos, animation and cartoon-style generation play a central role in creative industries, offering a structured yet expressive testbed for modeling motion, timing, exaggeration, and long-term character consistency.
Large-scale models have demonstrated strong capabilities in generating visually compelling motion, complex scenes, and extended temporal dynamics.
However, despite this rapid progress, the evaluation of animation video generation remains fundamentally under-developed.

Most existing video generation benchmarks~\cite{huang2024vbench, zheng2025vbench, huang2025vbench++, sun2025t2v, meng2024towards, wang2025your} are primarily designed for realistic or photorealistic videos, where evaluation emphasizes pixel fidelity, physical plausibility, and generic temporal consistency.
However, animation departs fundamentally from this regime: it relies on stylized visual abstractions, deliberately exaggerated motion, and character-centric acting that demands long-term identity and behavioral consistency.
As a result, evaluation criteria tailored for realism often fail to capture what defines animation quality, leading existing benchmarks to assign similar scores to videos that differ substantially in expressive motion, timing, and character performance.

Recent efforts~\cite{jiang2024anisora} attempt to adapt general video generation benchmarks to animation scenarios, but largely inherit evaluation dimensions from realistic video settings, focusing on overall visual quality or prompt adherence.
Such adaptations leave two fundamental gaps unresolved: (i) the absence of a principled definition of animation quality tailored to generative models, and (ii) the lack of systematic metrics for animation-specific properties, including expressive motion, character acting, and controlled exaggeration.
In professional animation practice, these aspects are governed by the \textit{Twelve Basic Principles of Animation}~\cite{johnston1981illusion}, which formalize how motion, timing, and acting convey weight, intent, and personality.
Despite their foundational role in animation theory and production, these principles have yet to be operationalized into a unified and model-agnostic evaluation framework for generative video models.

In this work, we introduce \textbf{AnimationBench}, a dedicated benchmark explicitly designed for evaluating animation video generation from a character-centric and motion-aware perspective.
AnimationBench is built upon three complementary pillars that jointly define animation quality for generative models.
First, we elevate \textit{IP Preservation} to a first-class evaluation target, assessing whether a model can maintain consistent character appearance, behavior, and personality across time—an essential requirement for character-driven animation rather than a by-product of visual similarity.
Second, we systematically operationalize the \textit{Twelve Basic Principles of Animation}~\cite{johnston1981illusion} into measurable evaluation dimensions, focusing on principles that govern perceptual motion quality and expressive acting, while avoiding subjective directorial choices.
Third, we incorporate \textit{Broader Quality Dimensions} that are critical for modern video models, including semantic consistency, motion rationality, and camera motion consistency, thereby complementing animation-specific evaluation with general generative video quality.
To the best of our knowledge, AnimationBench is the world’s first systematic animation benchmark. 

AnimationBench is designed as a flexible, extensible framework rather than a fixed benchmark. It offers two evaluation settings: a close-set setting with standardized prompts and protocols for reproducible comparisons, and an open-set setting for evaluating arbitrary animations, enabling diagnostic analysis and model refinement. This dual approach allows AnimationBench to serve as both a benchmark and a diagnostic tool. 
It leverages visual-language models (VLMs) for scalable and consistent evaluation.

By formulating animation assessment as structured visual reasoning tasks, our evaluation pipeline achieves strong alignment with human judgment while remaining extensible to new dimensions, protocols, and animation styles.
The main comparison between AnimationBench and other existing video benchmarks is illustrated in Table~\ref{tab:benchmark_comparison}, and a more detailed discussion of related work can be found in the Appendix. \ref{sec:related work}.


In summary, we make three main contributions:
\vspace{-3mm}
\begin{itemize}
    \item We introduce \textbf{AnimationBench}, the first benchmark explicitly designed for animation video generation, moving beyond realism-oriented evaluation by grounding animation quality in a principled, domain-aware framework based on professional animation theory. \vspace{-2mm}
    \item We elevate \textit{IP Preservation} to a first-class evaluation objective, providing systematic metrics to assess long-term consistency of character appearance, behavior, and personality—an essential yet previously under-explored requirement for character-centric animation generation. \vspace{-2mm}
    \item We propose a scalable, \textit{VLM-based} evaluation framework that supports both a standardized close-set protocol for reproducible model comparison and an open-set diagnostic setting for fine-grained analysis of arbitrary animation videos. \vspace{-2mm}
\end{itemize}

\begin{table}[t]
\centering
\caption{\textbf{Comparison of Video Generation Benchmarks from an Animation-Centric Perspective.}
We compare representative benchmarks by their evaluation standards and protocols.
\textbf{AnimationBench} is explicitly designed for animation video generation, incorporating animation-centric criteria, \emph{IP Preservation}, perceptual motion and acting evaluation, and supporting both \emph{close-set} and \emph{open-set} evaluation via a \emph{VLM-based} framework. Here ``/'' denotes \emph{not applicable} (the benchmark is not designed for character-centric animation); $^{*}$ indicates partial open-set support.}
\label{tab:benchmark_comparison}

\setlength{\tabcolsep}{3pt}
\renewcommand{\arraystretch}{0.95}

\resizebox{\columnwidth}{!}{
\begin{tabular}{l|ccc|cc}
\toprule
 & \multicolumn{3}{c|}{\textbf{Evaluation Standard}} 
 & \multicolumn{2}{c}{\textbf{Evaluation Protocol}} \\
\textbf{Benchmark} 
 & \textit{Animation-centric} 
 & \textit{IP Preservation} 
 & \textit{Motion / Acting} 
 & \textit{Open-set} 
 & \textit{VLM-based} \\
\midrule
VBench~\cite{huang2024vbench}              &               & /             &               & $\checkmark^{*}$              &               \\
T2V-CompBench~\cite{sun2025t2v}            &               & /             &               &               &               \\
PhyGenBench~\cite{meng2024towards}         &               & /             &               &               &               \\
StoryEval~\cite{wang2025your}              &               & /             &               &               &               \\
VBench-2.0~\cite{zheng2025vbench}          &               & /             &               & $\checkmark^{*}$ & $\checkmark$ \\
AniSora~\cite{jiang2024anisora}            & $\checkmark$  &               &               &               &               \\
\textbf{AnimationBench (Ours)}             & $\checkmark$  & $\checkmark$  & $\checkmark$  & $\checkmark$  & $\checkmark$ \\
\bottomrule
\end{tabular}
}
\vspace{-3mm}
\end{table}

\begin{figure*}[t!]
    \centering
    \includegraphics[width=\textwidth]{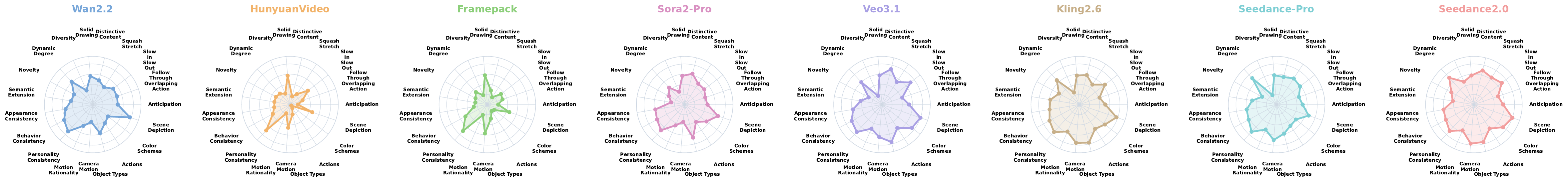}
    \caption{\textbf{Per-model dimension profiles on AnimationBench.}
    Each panel shows one model’s scores over the 19 dimensions (normalized per dimension); see Table~\ref{tab:overall} for the raw numbers.}
    \label{fig:each_score}
    \vspace{-3mm}
\end{figure*}

\section{AnimationBench Suite}
\label{sec:animationbench_suite}
This section describes the architectural components and evaluation frameworks of AnimationBench.
Our evaluation logic follows a top-down hierarchy:
\textbf{(i) IP Preservation} measures whether the generation content faithfully preserves the target IP identity;
\textbf{(ii) Animation Principles} measure whether the animation satisfies professional-quality motion and acting standards;
\textbf{(iii) Broader Quality Dimensions} capture general-purpose generation properties beyond animation-specific criteria, including semantic consistency, motion rationality, and camera motion consistency.

Except for a few dimensions evaluated with specialized evaluators, 
most dimensions requiring models with high-level understanding capabilities are assessed via \emph{VLM-based structured evaluation}.
Specifically, we formulate each dimension as a set of multi-question queries, 
where a salient concept is probed through multiple complementary yes/no questions, 
and the VLM’s responses are aggregated into a unified score.
Details are provided in the Appendix.~\ref{sec:VLM-based}.



\subsection{IP Preservation}
\label{sec:ip_preservation}
\textbf{IP preservation} is essential for evaluating animated content: in many successful animations, what audiences ultimately remember and follow is the IP itself---iconic characters, signature designs, and recognizable personalities. 

\begin{figure}[t!]
    \centering
    \includegraphics[width=\linewidth]{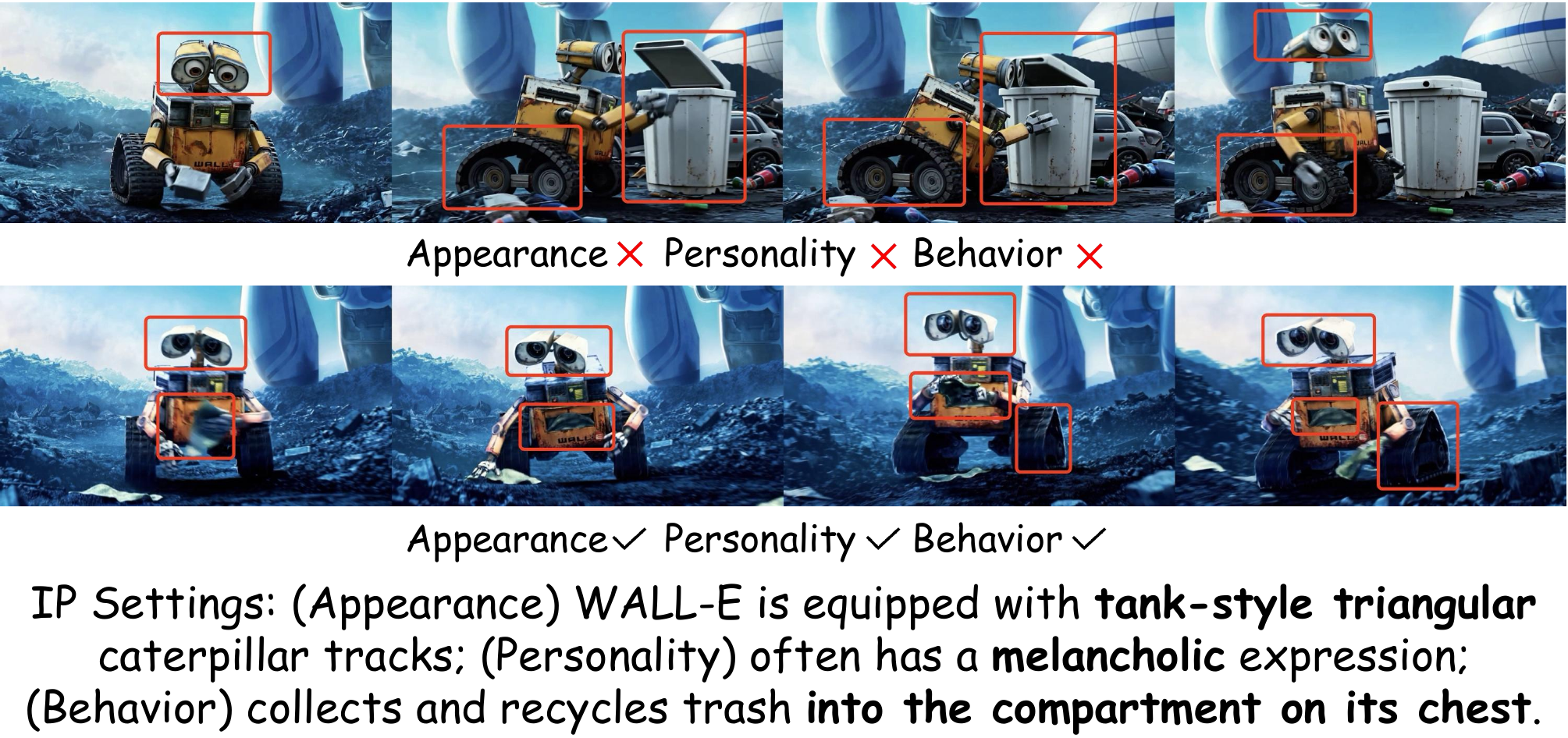}
    \caption{\textbf{IP Preservation Example.}
     \textbf{Up:} a failure case with noticeable IP drift generated by Seedance-Pro \cite{gao2025seedance}.
     \textbf{Down:} a success case with consistent appearance, behavior, and personality generated by Kling 2.6 \cite{kuaishou2024klingai}.}
    \label{fig:ip_consist}
    \vspace{-3mm}
\end{figure}

Different from generic character consistency, IP preservation must capture a broader notion of \emph{canonical identity} defined by the IP itself, spanning not only visual signatures but also characteristic behaviors and expressive performance.
For example, beyond keeping a character's outfit and proportions stable, an IP-faithful animation should depict a weapon-using character \emph{actually} fighting with their signature weapon rather than switching to bare-handed punches, and an evil character should not suddenly act overly righteous or display out-of-character emotions.
Motivated by these practical failure modes, we evaluate IP preservation from three complementary aspects (Fig.\ref{fig:ip_consist} illustrates an example):

\noindent\textbf{Appearance Consistency.} This category evaluates whether the character conforms to the canonical visual description of the target IP under viewpoint and motion changes. We consider two evaluation settings: (i) a dedicated 360$^\circ$ rotation setting, where the character is generated while turning to expose frontal, profile, and back views; and (ii) a general video setting, where we assess appearance stability under large-range motion (e.g., walking, running, or turning during an action). 

\noindent\textbf{Behavior Consistency.} This category assesses whether the character's actions align with established traits and narrative context. It covers (i) \emph{signature traits} such as weight, speed, gait, and characteristic motion style, (ii) \emph{environmental interaction}: how the character reacts to props and surroundings in a manner consistent with lore, and (iii) \emph{action logic}, the character's typical way of executing tasks or responding to situations.

\noindent\textbf{Personality Consistency.} This category evaluates whether the character's expressions and performance remain faithful to their intrinsic personality and emotional range. 
It focuses on \emph{facial acting} (e.g., eye movements, micro-expressions, reaction timing, and expression range) as well as broader \emph{personality cues} reflected through posture, attitude, presence, and body language typical of the IP.

Each aspect is evaluated via VLM-based yes/no questions, e.g., ``Does the character have a large round head with a dorsal fin?'' (appearance), ``Does the character use one claw to dig and use the other claw to pick it up?'' (behavior), and ``Is this character's expression cute, showing a cheerful personality?'' (personality).

\begin{table*}[t]
\centering
\caption{\textbf{AnimationBench Overall Evaluation Results.}
We report results following our top-down hierarchy: \textbf{IP Preservation}, \textbf{Animation Principles} (grouped into Motion Dynamics, Deformation, Expressiveness, and Human Preference), and \textbf{Broader Quality Dimensions}.
\textbf{Abbrev.}: \textbf{Antic.}=Anticipation; \textbf{FTOA}=Follow Through and Overlapping Action; \textbf{SI/SO}=Slow In and Slow Out; \textbf{S\&S}=Squash and Stretch; \textbf{DC}=Distinctive Content; \textbf{SD}=Solid Drawing; \textbf{DD}=Dynamic Degree; \textbf{SE}=Semantic Extension; \textbf{Sem.}=Semantic Consistency; \textbf{MR}=Motion Rationality; \textbf{CMC}=Camera Motion Consistency.}
\label{tab:overall}
{\setlength{\tabcolsep}{2.2pt}
\renewcommand{\arraystretch}{1.05}
\scriptsize
\resizebox{\textwidth}{!}{
\begin{tabular}{l|ccc|cccccccccc|ccccccc}
\toprule
& \multicolumn{3}{c|}{\textbf{IP Preservation}} 
& \multicolumn{10}{c|}{\textbf{Animation Principles}} 
& \multicolumn{7}{c}{\textbf{Broader Quality Dimensions}} \\
\cmidrule(lr){2-4}\cmidrule(lr){5-14}\cmidrule(lr){15-21}
\textbf{Models}
& \textbf{App.} & \textbf{Beh.} & \textbf{Pers.}
& \multicolumn{3}{c}{\textbf{Motion Dynamics}}
& \multicolumn{1}{c}{\textbf{Deform.}}
& \multicolumn{3}{c}{\textbf{Expressiveness}}
& \multicolumn{3}{c|}{\textbf{Human Pref.}}
& \multicolumn{5}{c}{\textbf{Sem. Consistency}}
& \textbf{MR} & \textbf{CMC} \\
\cmidrule(lr){5-7}\cmidrule(lr){8-8}\cmidrule(lr){9-11}\cmidrule(lr){12-14}\cmidrule(lr){15-19}
& & &
& \textbf{Antic.} & \textbf{FTOA} & \textbf{SI/SO}
& \textbf{S\&S}
& \textbf{DC} & \textbf{SD} & \textbf{Nov.}
& \textbf{Div.} & \textbf{DD} & \textbf{SE}
& \textbf{Overall} & \textbf{Obj.} & \textbf{Act.} & \textbf{Color} & \textbf{Scene}
& & \\
\midrule
\textbf{Wan2.2} 
& \cellcolor{osfirst}\textbf{67.69} & \cellcolor{osfirst}\textbf{80.09} & \cellcolor{osfirst}\textbf{89.58} 
& \cellcolor{osfirst}\textbf{61.59} & \cellcolor{osfirst}\textbf{63.16} & \cellcolor{osfirst}\textbf{63.76} 
& \cellcolor{osfirst}\textbf{50.94} 
& \cellcolor{osfirst}\textbf{61.67} & \cellcolor{osthird}70.43 & \cellcolor{osfirst}\textbf{15.87} 
& \cellcolor{osfirst}\textbf{38.14} & \cellcolor{osfirst}\textbf{76.67} & \cellcolor{osfirst}\textbf{54.14} 
& \cellcolor{osfirst}\textbf{67.86} & \cellcolor{osfirst}\textbf{72.62} & \cellcolor{osfirst}\textbf{54.76} & \cellcolor{osfirst}\textbf{47.61} & \cellcolor{osfirst}\textbf{96.43} 
& \cellcolor{osfirst}\textbf{57.26} & \cellcolor{osthird}42.86 \\

\textbf{HunyuanVideo} 
& \cellcolor{ossecond}39.48 & \cellcolor{osthird}47.91 & \cellcolor{osthird}86.53 
& \cellcolor{osthird}19.70 & \cellcolor{osthird}32.41 & \cellcolor{ossecond}55.76 
& \cellcolor{ossecond}30.82 
& \cellcolor{osthird}20.83 & \cellcolor{ossecond}71.38 & \cellcolor{ossecond}11.91 
& \cellcolor{ossecond}29.48 & \cellcolor{osthird}36.67 & \cellcolor{ossecond}39.10 
& \cellcolor{osthird}24.20 & \cellcolor{osthird}25.00 & \cellcolor{osthird}10.26 & \cellcolor{osthird}3.85 & \cellcolor{ossecond}57.69 
& \cellcolor{osthird}23.42 & \cellcolor{ossecond}57.14 \\

\textbf{Framepack} 
& \cellcolor{osthird}33.85 & \cellcolor{ossecond}61.30 & \cellcolor{ossecond}88.33 
& \cellcolor{ossecond}26.81 & \cellcolor{ossecond}39.47 & \cellcolor{osthird}42.22 
& \cellcolor{osthird}21.37 
& \cellcolor{ossecond}35.83 & \cellcolor{osfirst}\textbf{72.36} & \cellcolor{osthird}9.91 
& \cellcolor{osthird}25.36 & \cellcolor{ossecond}41.67 & \cellcolor{osthird}30.07 
& \cellcolor{ossecond}33.33 & \cellcolor{ossecond}35.71 & \cellcolor{ossecond}19.05 & \cellcolor{ossecond}21.43 & \cellcolor{osthird}57.14 
& \cellcolor{ossecond}27.92 & \cellcolor{osfirst}\textbf{71.43} \\

\midrule

\textbf{Sora2-Pro} 
& 73.20 & 77.31 & 86.01 
& 56.14 & \cellcolor{csthird}53.51 & 61.20 
& 61.25
& \cellcolor{csthird}78.33 & 71.02 & \cellcolor{csthird}13.53 
& \cellcolor{cssecond}34.53 & 56.67 & 33.83 
& 73.36 & 83.89 & 51.11 & 67.78 & 86.67 
& 55.86 & 42.86 \\

\textbf{Veo3.1} 
& 72.02 & \cellcolor{csfirst}\textbf{84.72} & \cellcolor{csfirst}\textbf{91.61} 
& \cellcolor{cssecond}66.67 & 52.63 & \cellcolor{csfirst}\textbf{88.80} 
& \cellcolor{csthird}66.48
& \cellcolor{csfirst}\textbf{90.00} & \cellcolor{cssecond}72.07 & 11.43 
& 23.40 & 75.00 & \cellcolor{csthird}61.25
& \cellcolor{csfirst}\textbf{89.31} & \cellcolor{cssecond}95.00 & \cellcolor{csthird}68.89 & \cellcolor{csfirst}\textbf{93.33} & \cellcolor{csfirst}\textbf{100.00} 
& 64.76 & 80.36 \\

\textbf{Kling2.6} 
& \cellcolor{csthird}74.09 & \cellcolor{cssecond}83.33 & \cellcolor{csfirst}\textbf{91.61} 
& \cellcolor{csthird}64.04 & 52.63 & \cellcolor{csthird}80.00 
& 58.74
& 75.00 & 71.81 & \cellcolor{cssecond}18.31 
& \cellcolor{csthird}32.48 & \cellcolor{csthird}81.67 & \cellcolor{csfirst}\textbf{72.18} 
& \cellcolor{csthird}86.11 & \cellcolor{csfirst}\textbf{96.67} & \cellcolor{csfirst}\textbf{71.11} & \cellcolor{csthird}80.00 & \cellcolor{csthird}96.67 
& \cellcolor{csfirst}\textbf{72.64} & \cellcolor{cssecond}94.64 \\

\textbf{Seedance-Pro} 
& \cellcolor{cssecond}74.61 & 78.24 & \cellcolor{csthird}90.91 
& \cellcolor{csthird}64.04 & \cellcolor{cssecond}57.90 & 73.20 
& \cellcolor{cssecond}76.91 
& 70.00 & \cellcolor{csfirst}\textbf{72.54} & 12.50 
& 25.48 & \cellcolor{cssecond}88.33 & \cellcolor{cssecond}60.90 
& 69.86 & 73.89 & 62.22 & 60.00 & 83.33 
& \cellcolor{csthird}76.91 & \cellcolor{csthird}87.50 \\

\textbf{Seedance2.0} 
& \cellcolor{csfirst}\textbf{75.83} & \cellcolor{csthird}78.44 & 88.33 
& \cellcolor{csfirst}\textbf{72.02} & \cellcolor{csfirst}\textbf{62.69} & \cellcolor{cssecond}86.67 
& \cellcolor{csfirst}\textbf{79.54} 
& \cellcolor{cssecond}86.67 & \cellcolor{csthird}72.02 & \cellcolor{csfirst}\textbf{19.58} 
& \cellcolor{csfirst}\textbf{61.57} & \cellcolor{csfirst}\textbf{89.73} & 52.63 
& \cellcolor{cssecond}89.05 & \cellcolor{cssecond}95.00 & \cellcolor{cssecond}70.48 & \cellcolor{cssecond}90.73 & \cellcolor{csfirst}\textbf{100.00} 
& \cellcolor{cssecond}69.43 & \cellcolor{csfirst}\textbf{96.42} \\

\bottomrule
\end{tabular}
}}
\vspace{-3mm}
\end{table*}

\subsection{Animation Principles}
\label{sec:TBPA}
Evaluating generative animation requires more than assessing generic video fidelity: high-quality animation is judged by \emph{performance}, whether motion and acting convey intent, emotion, and believability in a way that aligns with established production practice.
To ground our benchmark in an industry-recognized standard, we adopt the \textit{Twelve Principles of Animation}~\cite{johnston1981illusion}, which have long served as foundational guidelines in professional animation pipelines across film, television, and games. 
However, not all twelve principles are equally suitable as \emph{explicit} benchmark dimensions for generated animation video.
Some principles are inherently subjective or depend primarily on directorial intent rather than intrinsic animation quality in the rendered motion.
For example, \textbf{Staging} concerns shot composition and guiding viewer attention, which is largely determined by cinematography choices.
\textbf{Straight Ahead Action and Pose-to-Pose} describes animation \emph{workflows} (the authoring process) rather than perceptual properties directly observable in the generated video.
\textbf{Timing} is also highly context- and style-dependent, and is often inseparable from narrative intent and acting choices.

More broadly, even among the remaining principles, many are not independent when used for benchmarking: different principles often contribute to the same overarching goal of \emph{motion plausibility} and action coherence.
For example, while \textbf{Anticipation} and \textbf{Follow-through/Overlapping Action} emphasize different parts of an action, they both serve to make the motion feel physically grounded and reasonable to viewers.
Thus, instead of treating each principle as an isolated benchmark axis, we group principles that evaluate the same underlying quality into a unified dimension.
With these exclusions in place, we reorganize the remaining principles into four primary dimensions:
\textbf{Motion Dynamics}, \textbf{Deformation}, \textbf{Expressiveness}, and \textbf{Human Preference}.

\subsubsection{Motion Dynamics}
\textbf{Motion Dynamics} focuses on the physical and temporal coherence of motion, ensuring that the actions in the generated video are logically structured, with appropriate inertia and smooth transitions. This dimension includes principles such as \textbf{Anticipation}, \textbf{Follow Through and Overlapping Action}, \textbf{Slow In and Slow Out}, and \textbf{Arc \& Secondary Action}, all of which ensure that motion feels natural, fluid, and coherent.

\paragraph{Anticipation.}
This principle refers to a preparatory motion that signals an upcoming primary action, helping make the main action more readable and realistic. 
We evaluate whether the generated video includes a visually recognizable preparatory action before the main action by posing structured yes/no questions to a VLM.
For example: ``Does the character dip the brush into paint before beginning to paint?'' or ``Does the character crouch before jumping?''

\paragraph{Follow Through and Overlapping Action.}
This principle deals with inertia and delayed motion, where secondary parts (e.g., hair, clothing, accessories) continue moving after the main body slows or stops. 
We evaluate whether the motion transitions in the video exhibit a natural \textbf{follow-through} pattern (e.g., gradual deceleration and residual motion) and whether secondary elements exhibit \textbf{overlapping} motion that lags behind the primary movement, again using VLM-based structured QA.
Example questions include: ``Does the character’s hair continue moving after the body has stopped?'' and ``Does the character’s clothing lag behind the body movement as expected?''

\paragraph{Slow In and Slow Out.}
This principle states that motion typically starts with gradual acceleration, reaches peak speed in the middle, and decelerates smoothly toward the end, mimicking real-world inertia to avoid robotic or abrupt movement.

To evaluate this, we extract the trajectory of the primary moving object from the generated video using CoTracker~\cite{karaev2024cotracker}, and compute a frame-wise relative motion speed from foreground and background motion. The speed curve is smoothed using a moving average with window size 9, and the detected motion interval is divided into the first 20\%, the middle part, and the last 20\%. We then apply an explicit numerical heuristic to determine whether the speed profile follows the expected pattern of gradual acceleration, a clear mid-motion peak, and gradual deceleration. Specifically, we require the peak-to-valley ratio to be at least 2.0, and both the start-to-middle increase and the middle-to-end decrease to satisfy absolute and relative thresholds of 0.15 and 20\%, respectively. These criteria contribute up to 3 points. In addition, both the acceleration and deceleration phases are required to last at least 5\% of the video length, contributing 2 additional points, for a total score ranging from 0 to 5.

\paragraph{Arc \& Secondary Action.}
\textbf{Arc} encourages curved, natural trajectories, while \textbf{Secondary Action} adds supportive motions that enhance the main action.
Since both are highly context-dependent and hard to disentangle as standalone criteria in generated videos, we follow their intended objective and capture their effects within our Sec.\ref{sec:MR} \textbf{Motion Rationality} evaluation.

\subsubsection{Deformation}
\begin{figure}[t!]
    \centering
    \includegraphics[width=\linewidth]{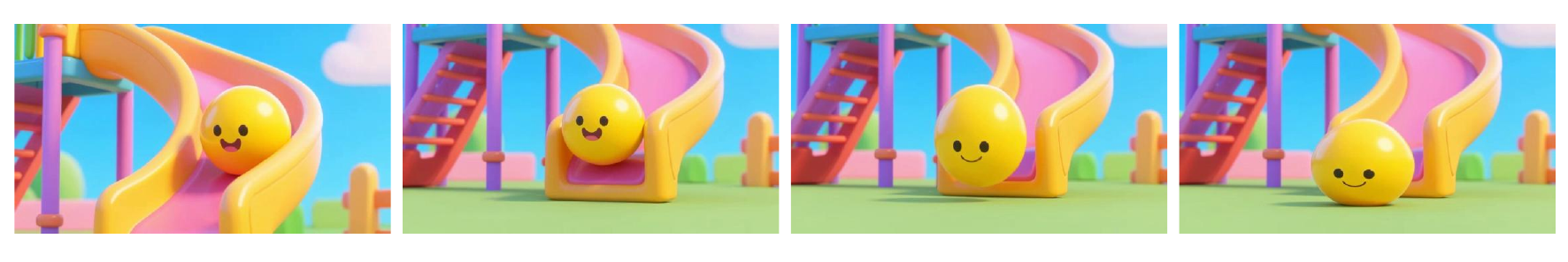}
    \caption{\textbf{Squash and Stretch Example.} We show one successful case (generated by Seedance-Pro \cite{gao2025seedance}) that exhibit clear squash-and-stretch deformation.}
    \label{fig:cartoon ball}
    \vspace{-3mm}
\end{figure}

\textbf{Deformation} refers to how objects and characters change shape during motion, ensuring that exaggeration is appropriately applied to maintain visual interest and physical plausibility.
To evaluate this dimension, we focus on the \textbf{Squash and Stretch} principle, which captures whether shape changes convey a sense of weight and flexibility through controlled exaggeration, such as flattening at impact and elongating during rebound (see Fig.~\ref{fig:cartoon ball} as an example).

Since true 3D volume is difficult to recover from generated videos, we evaluate this property in 2D using a combination of \textbf{area preservation} and \textbf{deformation magnitude}. We first ask a VLM whether a \emph{rebound} occurs; if not, the score is set to $0$. Otherwise, we apply a video segmentation model~\cite{carion2025sam3segmentconcepts} to track the target object and obtain a binary mask $\mathcal{M}_t$ for each frame $t$, with area $A_t = |\mathcal{M}_t|$.

For area preservation, we compute:
\begin{equation}
\begin{aligned}
    r_t &= \frac{|A_t - A_{t-1}|}{A_{t-1} + \epsilon}, \\
    \bar{r} &= \frac{1}{T-1}\sum_{t=2}^{T} r_t, \\
    S &= 100 \cdot \left(1 - \min(1, \bar{r})\right),
\end{aligned}
\end{equation}
where $r_t$ is the per-frame area change rate, $\bar{r}$ is the mean area variation across the sequence, $T$ is the total number of frames, and $\epsilon$ is a small constant to avoid division by zero. A higher $S$ indicates that the object better preserves its apparent area during deformation.

To explicitly capture visible squash-and-stretch, we further measure shape deformation from the mask geometry. For each frame $t$, given $\mathcal{M}_t$, we compute the covariance matrix of the foreground pixels and obtain its two eigenvalues $\lambda_{t,1} \geq \lambda_{t,2}$. We then define the shape anisotropy descriptor as
\begin{equation}
    u_t = \log \frac{\sqrt{\lambda_{t,1}} + \epsilon}{\sqrt{\lambda_{t,2}} + \epsilon}.
\end{equation}
This quantity measures how elongated or compressed the object appears at frame $t$. Based on this descriptor, we define the temporal deformation magnitude as
\begin{equation}
\begin{aligned}
    d_t &= |u_t - u_{t-1}|, \\
    \bar{d} &= \frac{1}{T-1}\sum_{t=2}^{T} d_t, \\
    D &= 100 \cdot \min\left(1, \frac{\bar{d}}{\tau}\right),
\end{aligned}
\end{equation}
where $\tau$ is a normalization constant controlling the saturation of the deformation reward. A higher $D$ indicates stronger visible shape change over time.

Finally, we combine the two terms to obtain the squash-and-stretch score:
\begin{equation}
    W_2 =
    \begin{cases}
        0, & \text{if no rebound is detected},\\
        0.7S + 0.3D, & \text{otherwise}.
    \end{cases}
\end{equation}

In this way, the metric jointly evaluates whether the object maintains plausible area preservation while also exhibiting visible squash-and-stretch deformation during rebound.

\subsubsection{Expressiveness}
\textbf{Expressiveness} measures how effectively the animation conveys the character's intent and emotion through visual cues. It includes principles like \textbf{Exaggeration} and \textbf{Solid Drawing}.

\paragraph{Distinctiveness and Novelty.} Exaggeration is a hallmark of animation: rather than perfectly imitating reality, it selectively amplifies movements, expressions, or events to make actions clearer, more dynamic, and stylistically expressive. The appropriate degree of exaggeration depends on the style of the animation, but the core idea is to stay semantically faithful while presenting a more vivid or extreme rendition that avoids looking static or dull.
We borrow this idea and evaluate exaggeration-related capabilities from two complementary perspectives: \textbf{Distinctive Content} and \textbf{Novelty}.
\begin{figure}[H]
    \centering
    \includegraphics[width=\linewidth]{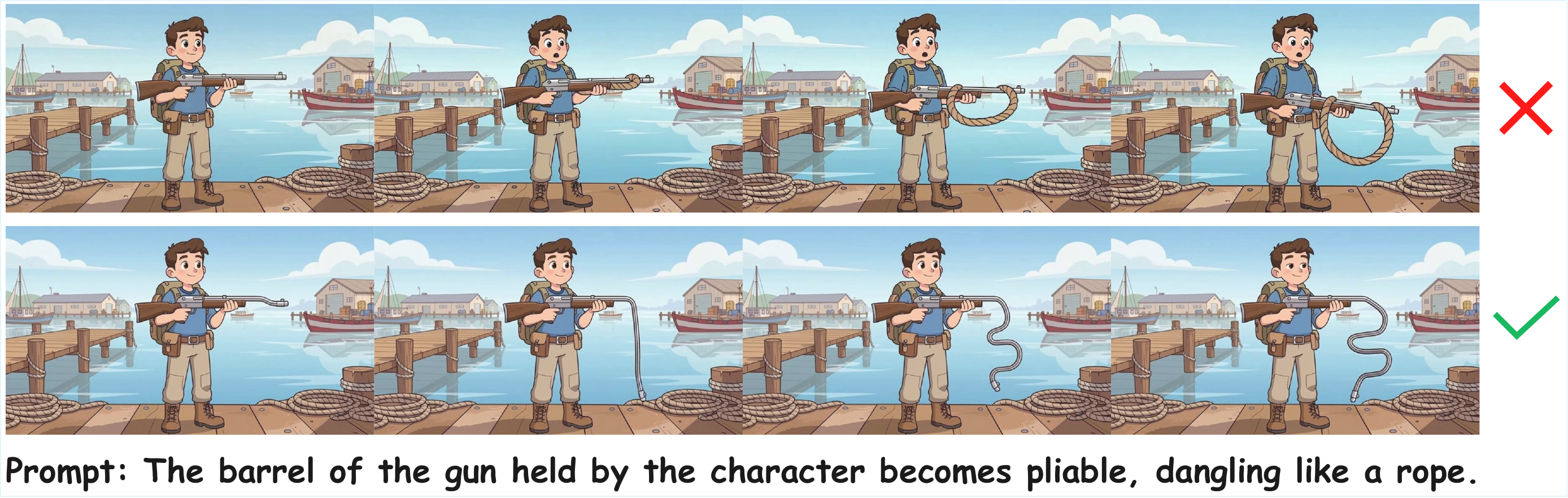}
    \caption{\textbf{Distinctive Content Example.} \textbf{Up:} failure case generated by Seedance-Pro \cite{gao2025seedance}. \textbf{Down:} success case generated by Sora2-Pro~\cite{openai2024sora}.}
    \label{fig:distinctive_content}
    \vspace{-3mm}
\end{figure}
\begin{itemize}
    \item \textbf{Distinctive Content.} We assess whether the model can generate novel or uncommon elements that are characteristic of animation, while still staying true to the prompt (see Fig.\ref{fig:distinctive_content} as an example). 
    The questions look like "Does the animation show [specific action or feature]?" or "Is the character performing [specific movement]?"
    \item \textbf{Novelty.} Novelty measures how much the outputs deviate from common patterns, thereby reflecting their ability to go beyond predictable animation behavior. For each test case, we construct a case-specific reference video that shares the same input image, and thus the same character and scene setup, but depicts a relatively normal or minimally extended animation. The reference video is generated separately using a model outside the tested set and is manually screened for suitability. We then extract video features using V-JEPA2~\cite{assran2025v} and compute the cosine similarity between the generated video and its paired reference video. Novelty is defined as $1 - \mu$, where $\mu$ denotes the mean similarity, with higher scores indicating greater deviation from the baseline and thus higher novelty.

\end{itemize}

\textbf{Solid Drawing} concerns the convincing depiction of three-dimensional form, volume, and weight. In image-to-video generation, these properties are largely inherited from the input image rather than newly created by the video model. Therefore, we do not assess Solid Drawing solely through a dedicated geometry metric. Instead, related aspects are reflected across multiple dimensions: Motion Rationality captures structural coherence during motion, IP Preservation reflects the stability of perceived form across frames and views, and Deformation measures volume and area consistency under shape changes. In addition, we use MUSIQ~\cite{ke2021musiq} as a perceptual quality proxy to capture low-level artifacts, such as overexposure, noise, or blur, that may weaken the perceived solidity of the generated result.

\subsubsection{Human Preference}
\textbf{Human Preference} evaluates the overall aesthetic and engagement quality of the animation, focusing on how appealing, captivating, and engaging the animation is. 

\begin{figure}[t]
    \centering
    \includegraphics[width=\linewidth]{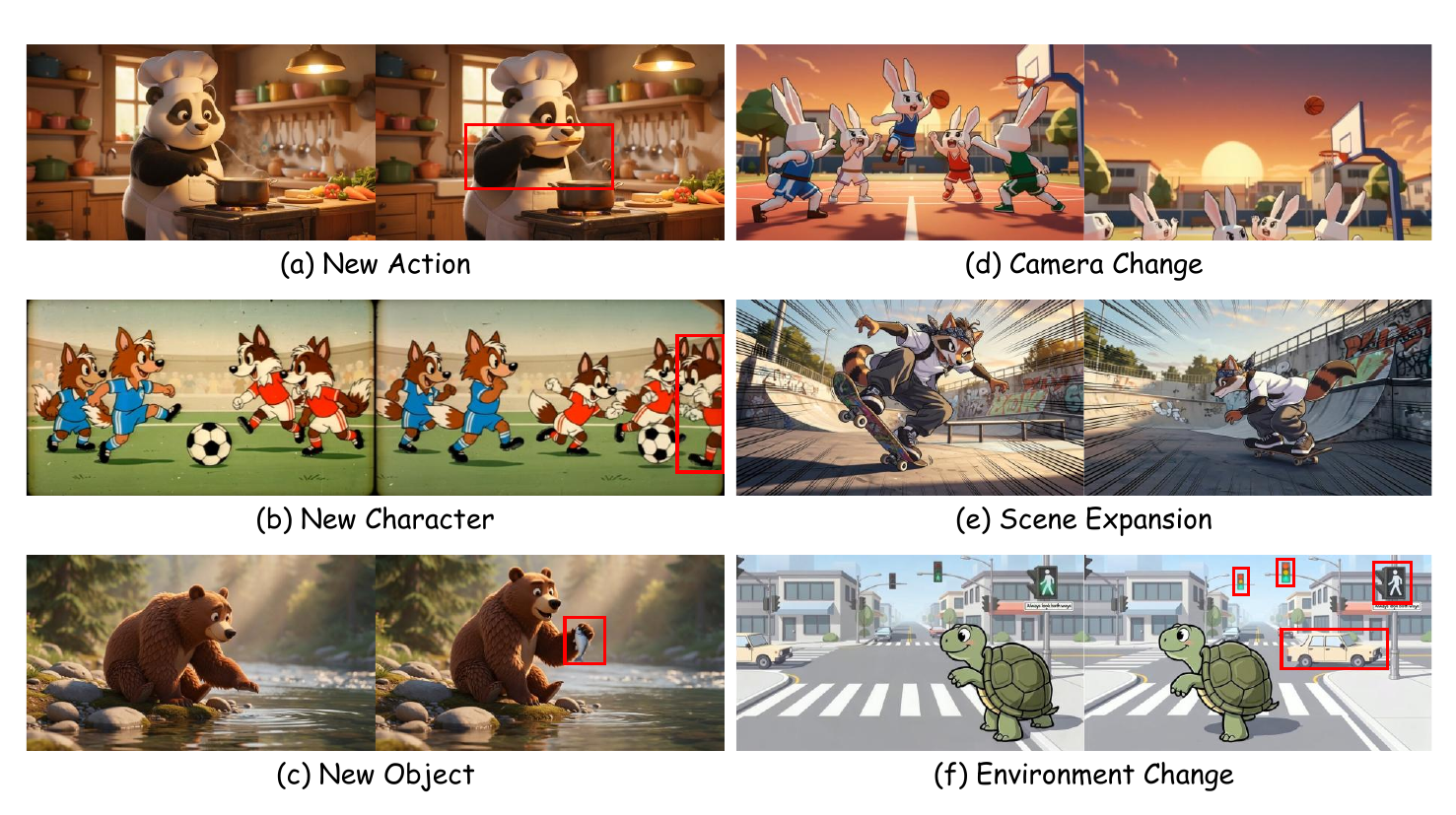}
    \caption{\textbf{Semantic Extension evaluation.} Assesses whether a model generates semantically plausible extensions beyond the prompt across six aspects: (a) new actions, (b) new characters, (c) new objects or interactions, (d) camera/editing changes, (e) coherent scene expansion, and (f) environmental changes}
    \label{fig:semantic_extension}
    \vspace{-3mm}
\end{figure}

To evaluate this dimension, we focus on the \textbf{Appeal} principle. 
Appeal is not about being ``nice'' or sympathetic (villains can be appealing too), but about the visual charm and engagement that make the animation interesting and worth watching.
To operationalize appeal, we use three measurable proxies: (i) motion activity, (ii) diversity of generated variants, and (iii) semantic extension.
So we use the following subdimensions:
\begin{itemize}
    \item \textbf{Dynamic Degree} evaluates the motion activity in the video to ensure it does not appear static. Following the implementation in VBench~\cite{huang2024vbench}, we use RAFT~\cite{teed2020raft} to estimate optical flow between adjacent frames and define the motion score of each frame pair as the mean magnitude of the top 5\% largest flow vectors. A frame pair is considered ``moving'' if this score exceeds $6.0 \times \frac{\min(H, W)}{256}$, where $H$ and $W$ denote the video height and width. At the video level, we sample frames at approximately 8 FPS and classify a video as ``dynamic'' if the number of moving frame pairs reaches the VBench~\cite{huang2024vbench} threshold, which is approximately equivalent to requiring motion in 25\% of the sampled temporal steps.
    
    \item \textbf{Diversity} measures the variation in generated samples from the same input. We sample five videos and compute the style/content diversity using features from the pre-trained VGG-19 model \cite{simonyan2014very}.

    \item \textbf{Semantic Extension.} As shown in Fig.~\ref{fig:semantic_extension}, we evaluate whether the model can generate semantically reasonable extensions beyond what is explicitly stated in the prompt. For this, we assess six aspects: (1) new actions, (2) new characters, (3) new objects or interactions, (4) camera/editing changes (e.g., cuts, angles, movement), (5) coherent scene expansion, and (6) environmental changes (e.g., lighting, weather, objects). New actions are counted individually, while other aspects are scored as yes = 1, no = 0. 
\end{itemize}
\subsection{Broader Quality Dimensions} \label{sec:generation related}
\subsubsection{Semantic Consistency}
Semantic consistency measures how well the generated content aligns with the input conditions in terms of meaning without considering aspects related to video quality.

Following VBench~\cite{huang2024vbench}, we evaluate semantic consistency along four categories: \textbf{object types}, \textbf{actions}, \textbf{color schemes}, and \textbf{scene depiction}. 
For each category, we check whether the generated video matches the input conditions, including object/style consistency with the prompt and conditioning image, the presence of prompted actions (without judging action quality), temporal color consistency, and alignment of background/scene with the described environment and style.


\begin{figure*}[t!]
    \centering
    \includegraphics[width=\textwidth]{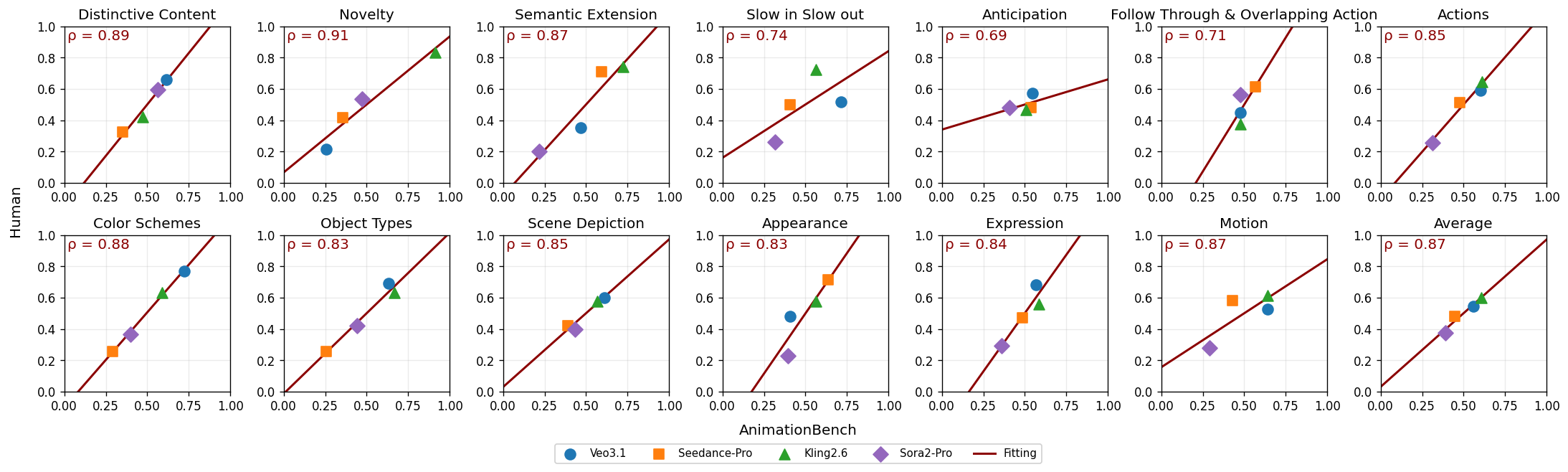}
    \caption{\textbf{Human alignment of AnimationBench.}
    We validate that AnimationBench scores are consistent with human preferences across dimensions.
    Each panel corresponds to one AnimationBench dimension: each dot is a model, with the AnimationBench win-rate on the x-axis and the human win-rate on the y-axis.
    We fit a linear trend line for visualization and report Spearman’s rank correlation coefficient ($\rho$) for each dimension.}
    \label{fig:human_alignment}
    \vspace{-6mm}
\end{figure*}

\subsubsection{Motion Rationality}
\label{sec:MR}

\textbf{Motion Rationality} evaluates whether the generated video exhibits coherent, natural, and context-appropriate motion.
It focuses on action-level plausibility and interaction logic, assessing whether the performed actions are fluent and reasonable, and whether the character or object behaves consistently with the prompt and the scene context.

Specifically, this dimension covers:
(i) \emph{structural integrity} (e.g., absence of extra limbs or physically impossible artifacts);
(ii) \emph{motion coherence} (e.g., temporally continuous, non-jerky motion rather than teleportation-like jumps);
(iii) \emph{action semantic plausibility} (e.g., ``swimming'' is realized with swimming-like movements and appropriate displacement); and
(iv) \emph{environmental interaction consistency} (e.g., actions logically fit the environment depicted by the conditioning image and the prompt).


\subsubsection{Camera Motion Consistency}
Camera Motion Consistency assesses whether the generated video exhibits motion patterns that are consistent with the intended camera operation described in the prompt. Specifically, it evaluates if the observed scene dynamics correctly correspond to the commanded camera control, such as panning, tilting, or zooming.
We provide paired text prompts for different camera motions along with corresponding images to drive generation. We focus on seven motion types: ``pan left'', ``pan right'', ``tilt up'', ``tilt down'', ``zoom in'', ``zoom out'', and ``static''. We use the Co-Tracker \cite{karaev2024cotracker} to track points along the four edges of the video and predict the camera motion type using a carefully designed algorithm following VBench++ \cite{huang2025vbench++}.

\subsection{Open-Set Refinement Pipeline}
\label{sec:openset}
Our open-set refinement pipeline is designed to diagnose issues in an arbitrary animation video with respect to a target evaluation dimension, and to further improve generation quality via prompt refinement.
Given an input video $V$ and an original prompt $T$, we provide the target evaluation dimension as a system prompt to a VLM.
Conditioned on this dimension specification, the VLM identifies potential failure modes and automatically generates a set of diagnostic questions that highlight what is missing or inconsistent in $V$ under the chosen dimension.
These questions, together with the original prompt $T$, are then fed into a prompt refiner, which produces an improved prompt $\hat{T}$ that better enforces the desired behaviors.
Finally, the refined prompt $\hat{T}$ is used to re-run the video generation model, yielding a higher-quality video that better satisfies the selected evaluation dimension.

This differs from our close-set setting, where prompts and questions are pre-constructed and fixed. 
In the open-set setting, the VLM dynamically derives dimension-specific questions for each input video to support targeted diagnosis and refinement.

\section{Experiments}
We evaluate seven state-of-the-art video generation models, including open-source ones: Wan2.2~\cite{wan2025wan}, HunyuanVideo~\cite{wu2025hunyuanvideo}, and Framepack~\cite{zhang2025packing}, as well as closed-source ones: Sora2-Pro~\cite{openai2024sora}, Veo3.1~\cite{veo2025}, Kling2.6~\cite{kuaishou2024klingai}, Seedance-Pro~\cite{gao2025seedance} and Seedance2.0~\cite{bytedance_seed_seedance2_2026}. 
Model-specific information is provided in Appendix.~\ref{sec:model sum}.

\begin{figure}[t!]
    \centering
    \includegraphics[width=\linewidth]{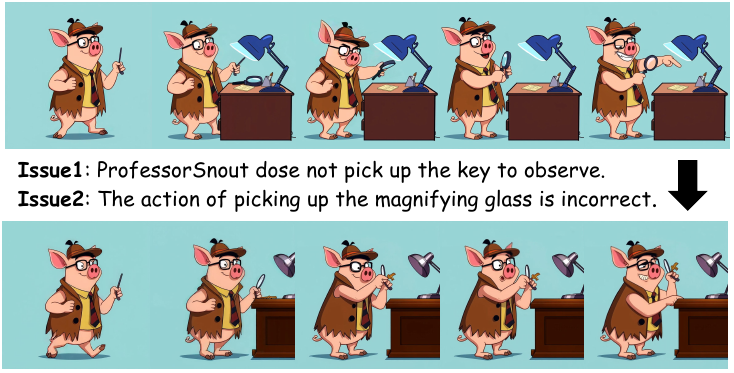}
    \caption{\textbf{Open-set refined example.} After open-set refinement, the video effect of ProfessorSnout performing clue inspection has been significantly improved.}
    \label{fig:openset_case2}
    \vspace{-8mm}
\end{figure}

For each dimension of evaluation, we compute the AnimationBench scores using the evaluation tools described in Section~\ref{sec:animationbench_suite}, and report the results in Fig.~\ref{fig:overall score}, Fig.~\ref{fig:each_score} and Table~\ref{tab:overall}. 
For the open-set category, we applied the method described in Section~\ref{sec:animationbench_suite}. Specifically, we optimized Wan2.2 \cite{wan2025wan} using the same image suite used for semantic consistency. As shown in Fig.~\ref{fig:openset_case2}, this approach fixed the majority of the issues, leading to a significant improvement in the semantic consistency of the generated videos.

\subsection{Evaluation Setting}


For evaluation dimensions about  IP consistency, we build a cartoon character set consisting of both collected existing IPs and self-designed IPs. Specifically, we collect \textbf{10 existing IPs} with character profiles, reference images, and multi-view sheets, and further create \textbf{30 self-designed IPs} to ensure a fairer evaluation by avoiding potential training bias of video models for popular characters. Overall, this set contains \textbf{30 2D and 10 3D} characters spanning diverse categories (e.g., animals, robots, humans) and styles (e.g., Disney, Japanese anime, minimalism, American comics, 1990s retro). We use Qwen-Image-Edit~\cite{wu2025qwen} to synthesize starting frames by editing backgrounds and initial poses to match the prompts.
For evaluation dimensions that do not require specific IPs, such as distinctive content and novelty, we use a large language model to generate a variety of diverse cartoon character descriptions. Based on these descriptions, we then create corresponding input images adapted to different artistic styles and video prompts. 
Ultimately, based on a set of \textbf{170 source images} and \textbf{360 customized prompts}, each model is required to generate a total of \textbf{360 videos}. The specific distribution of these videos across different dimensions is detailed in Table~\ref{tab:number_of_generate}.
For the VLM backbone, after selecting the most reliable evaluator through a targeted diagnostic experiment (see Appendix.~\ref{sec:VLM-based}), we adopt Qwen3-VL-MAX~\cite{Qwen3-VL}. and use Qwen3-MAX \cite{qwen3} as our prompt refiner.


\subsection{Human Alignment of AnimationBench}
To verify that our automatic evaluation aligns with human perception, we conducted a human preference study on videos from four closed-source models. Annotators were shown paired model outputs for each dimension and selected the better one based on the criterion, with a tie option for comparable quality. We quantified human preference using a win-ratio metric (1 for wins, 0 for losses, 0.5 for ties). The per-dimension and overall correlations between the automatic AnimationBench scores and human win ratios, shown in Fig.~\ref{fig:human_alignment}, confirm the reliability of our evaluation framework. Spearman’s correlation coefficient was used to measure the consistency between AnimationBench’s per-dimension evaluations and human judgments. More details on annotator background, annotation procedure, and agreement analysis are provided in Appendix.~\ref{appendix:annotator-details}.

\subsection{Insights and Discussions}
\textbf{Overall Performance Evaluation:}
In terms of overall performance, the closed-source models, particularly Kling2.6 \cite{kuaishou2024klingai} Veo3.1 \cite{veo2025} and Seedance2.0~\cite{bytedance_seed_seedance2_2026}, demonstrate the best results.  
For the open-source models, Wan2.2 \cite{wan2025wan} shows notable capabilities, particularly excelling in some dimensions, rivaling even the closed-source models in certain areas. 
Overall, the models perform well in terms of visual fidelity and high-level semantic understanding. 
However, challenges remain in areas like expressiveness and IP consistency. 
While models can generate convincing visuals, maintaining true IP characteristics, handling exaggerated expressions, and preserving motion accuracy remain significant hurdles.

\textbf{Detailed Dimension Evaluation:}
When diving into more granular dimensions, certain patterns emerge:
\textit{Deformation:} Models like Sora2-Pro \cite{openai2024sora} and Kling2.6 \cite{kuaishou2024klingai} demonstrate deformation capabilities, but there's still a significant gap when compared to realistic deformations.
\textit{360-Degree Consistency:} Most models struggle to maintain consistent appearance from all angles.
Many (like HunyuanVideo \cite{wu2025hunyuanvideo} and Framepack \cite{zhang2025packing}) fail to generate 360-degree views.
\textit{IP Expression Control:} Managing IP-specific facial expressions remains a weak point for most models. Exaggerated facial expressions are particularly prone to errors, reflecting the difficulty of capturing the unique expression style of certain IPs.

\textbf{Significance of New Dimensions:}
Through this work, we have identified the importance and necessity of dimensions such as \textit{semantic extension} and \textit{anticipation}, which were previously overlooked.
These dimensions are crucial for evaluating a model’s ability to generate diverse, creative, and contextually consistent animations, offering a more comprehensive measure of performance beyond basic visual fidelity.

In summary, while overall performance across models has made substantial strides, there's still work to be done in achieving consistent, high-quality motion, true IP preservation, and nuanced expressiveness. The models are progressing, but these finer dimensions of animation quality are critical for future advancements.

\section{Conclusion}
As video generation models are increasingly used for animation creation, evaluating their animation-specific capabilities becomes both necessary and non-trivial.
In this work, we introduce \textbf{AnimationBench}, a dedicated benchmark tailored to animation image-to-video generation.
AnimationBench grounds its evaluation in \textbf{IP Preservation} and the \textbf{Twelve Basic Principles of Animation}, and supports both close-set benchmarking and open-set diagnosis through a mainly VLM-centric pipeline.
This design enables multi-dimensional, human-aligned, and actionable assessment of animation quality.
We hope AnimationBench will help the community measure progress more reliably and drive future advances toward stronger motion, acting, and identity-faithful animation generation.

\newpage
\section*{Impact Statements}
This paper aims to advance the evaluation of animation quality in video generation models. The benchmark is intended for research and evaluation only.

To ensure ethical and legal compliance, copyrighted materials are not directly redistributed in the dataset. When relevant cases are needed, we use references to official sources or replace them with non-copyrighted or properly licensed alternatives. We also emphasize that all data collection and release should follow applicable copyright rules, licensing requirements, and institutional policies.


\bibliography{example_paper}
\bibliographystyle{icml2026}
\newpage
\appendix
\newpage
\section{Related Work} \label{sec:related work}
\subsection{Video Generation Benchmark and Evaluation}
Evaluating video generation models has been a persistent challenge in recent years. Benchmarks such as the VBench series~\citep{huang2024vbench,huang2025vbench++,zheng2025vbench} have been developed for general video generation, assessing aspects from pixel fidelity to more complex intrinsic qualities. However, these metrics are primarily grounded in real-world video characteristics and do not adequately capture the stylized and expressive nature of animation. While efforts like the AniSora~\citep{jiang2024anisora} benchmark have introduced relevant criteria for animated content, existing frameworks remain largely adapted from conventional video evaluation and do not specifically address the unique demands and challenges of animation video generation.
\subsection{Animation Video Generation}
The development of diffusion models has driven significant progress in general video generation, enabling the synthesis of highly realistic and convincing videos~\citep{wan2025wan,wu2025hunyuanvideo,zhang2025packing,DBLP:journals/corr/abs-2503-09642}. However, the field of stylized animation video generation remains underdeveloped. This gap primarily stems from two challenges: a scarcity of large-scale, high-quality animation training data and the inherent stylistic and motion differences between animation and real-world videos. While some works have explored this direction~\citep{jiang2024anisora}, progress in animation video generation has not kept pace with advances in general video synthesis, underscoring the need for dedicated research efforts and evaluation frameworks tailored to the unique demands of animated content.
\subsection{Visual-Language Models in Video Evaluation}
The rise of powerful visual-language models (VLMs) has reshaped automated video evaluation, enabling the assessment of complex, semantic dimensions that align closely with human judgment. This shift is driven by the advanced multimodal capabilities of models such as Qwen3-VL~\citep{Qwen3-VL}, Gemini2.5-Pro~\citep{comanici2025gemini}, and GPT5~\citep{singh2025openai}. Recent benchmarks increasingly integrate VLMs to evaluate abstract qualities, such as stylistic consistency and narrative coherence, that are difficult to capture with conventional metrics~\citep{zheng2025vbench,he2025videoscore2,li2025worldmodelbench}. These works demonstrate the potential of VLMs in providing more nuanced, context-aware judgments, particularly in domains such as animation, where understanding style, expression, and artistic intent is essential.

\section{Model Specifications} \label{sec:model sum}
We specify the static properties of tested video models, as listed in Table~\ref{tab:video_model_comparison}.

\begin{table}[t]
\centering
\small
\setlength{\tabcolsep}{3.5pt}
\begin{tabular}{lccccc}
\toprule
\textbf{Model} & \textbf{IP} & \textbf{Motion} & \textbf{Pers.} & \textbf{Act./Exp.} & \textbf{Overall} \\
\midrule
Qwen2.5-VL-3B   & 67.5 & 61.0 & 56.5 & 63.0 & 62.0 \\
Qwen2.5-VL-72B  & 79.0 & 73.5 & 70.5 & 76.0 & 74.8 \\
InternVL3.5-14B & 75.5 & 69.5 & 67.0 & 72.0 & 71.0 \\
Qwen3-VL-Max    & \textbf{88.0} & \textbf{83.5} & \textbf{80.5} & \textbf{85.5} & \textbf{84.4} \\
\bottomrule
\end{tabular}
\caption{Evaluator-selection results on diagnostic cases with known failure patterns. Accuracy (\%) is computed based on whether the VLM correctly identifies the intended problem.}
\label{tab:vlm_selection}
\end{table}

\begin{table*}[t]
\centering
\small
\setlength{\tabcolsep}{6pt}
\begin{tabular}{lcccl}
\toprule
\textbf{Model} & 
\textbf{GPU Memory} & 
\textbf{Resolution / Settings} & 
\textbf{Params} & 
\textbf{Latency (Inf. / API)} \\
\midrule

FramePack~\cite{zhang2025packing} &
26.71\,GB &
640p, infinite, latent window=9 &
13B &
286\,s / 129 frames \\

Hunyuan Video~\cite{wu2025hunyuanvideo} &
Peak 60\,GB &
720p &
NA &
1667\,s / 129 frames \\

Wan 2.2~\cite{wan2025wan} &
45\,GB &
832$\times$480; 
720$\times$1280 / 1280$\times$720 &
14B &
NA \\


Seedance Pro~\cite{gao2025seedance} &
NA &
1080p &
NA &
1--2\,s \\

Seedance 2.0~\cite{bytedance_seed_seedance2_2026} &
NA &
720p &
NA &
$\sim$4\,min \\



Kling 2.6~\cite{kuaishou2024klingai} &
NA &
1080p &
NA &
$\sim$2\,min \\


VEO 3.1~\cite{veo2025} &
NA &
720p &
NA &
2--3\,min \\



Sora 2 Pro~\cite{openai2024sora} &
NA &
1080p, 20\,s video &
NA &
NA \\


\bottomrule
\end{tabular}
\caption{Comparison of representative video generation models in terms of GPU memory consumption, resolution settings, parameter scale, and inference latency. }
\label{tab:video_model_comparison}
\end{table*}

\section{Benchmark Details} \label{sec:VLM-based}
\paragraph{Selection of VLM.}
To determine the VLM evaluator used in our framework, we conducted a targeted evaluator-selection experiment on diagnostic cases with known failure patterns. Specifically, we measured whether each VLM could correctly recognize the intended problem in four dimensions: IP preservation, motion dynamics, personality, and acting/expression. The results are summarized in Table~\ref{tab:vlm_selection}. Qwen3-VL-Max~\cite{Qwen3-VL} achieves the best overall accuracy and consistently outperforms the other candidates across all four dimensions. Based on these results, we adopt Qwen3-VL-Max~\cite{Qwen3-VL} as the evaluator in all experiments.

\paragraph{VLM-based structured evaluation.}
Many of our dimensions are difficult to measure with purely low-level signals, yet they can be expressed as
\emph{structured} perceptual checks (e.g., whether a preparatory motion happens \emph{before} the main action, or whether an iconic outfit detail is preserved throughout).
We therefore formulate several dimensions as video question-answering (VQA) tasks verified by vision-language models (VLMs),
by constructing a set of questions $\mathcal{Q}=\{\mathcal{Q}^{(k)}\}_{k=1}^{K}$ that probe the same concept from different angles.
For example, for a dimension focusing on color change, we may ask:
(1) ``Initially, is the color of the river mostly blue?'';
(2) ``Finally, is the color of the river mostly brown?'';
(3) ``Does the color of the river change gradually over time rather than suddenly?''

\paragraph{Unified VLM-QA scoring.}
Given a video $V$ and a set of yes/no questions $\{\mathcal{Q}^{(k)}\}_{k=1}^{K}$,
we map each answer to $a_k\in\{0,1\}$ (Yes=1, No=0) and compute
\begin{equation}
\label{eq:vlm_qa_score}
S_{\mathrm{QA}}(\mathcal{Q},V) = \frac{100}{K}\sum_{k=1}^{K} a_k \in [0,100].
\end{equation}
We reuse Eq.~\ref{eq:vlm_qa_score} across dimensions to avoid repeated formulas.

\begin{figure}[t]
    \centering
    \includegraphics[width=\linewidth]{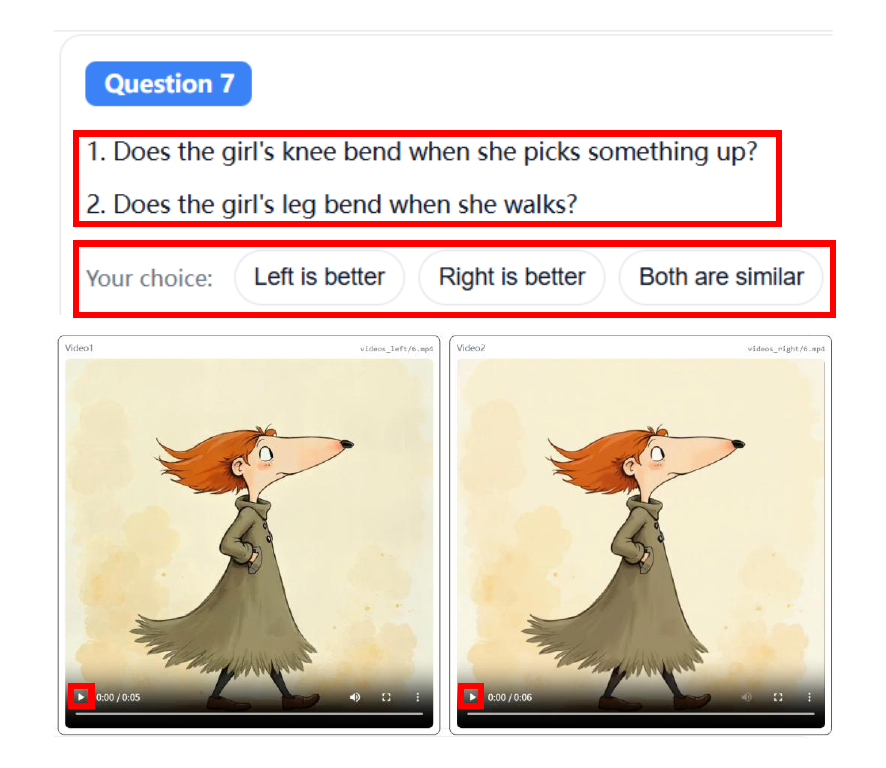}
    \caption{\textbf{Human Preference Annotation Interface}. Given the questions and the paired videos, users select the preferred option or indicate a tie.}
    \label{fig:user study}
\end{figure}

\begin{figure}[t]
\centering
\includegraphics[width=\linewidth]{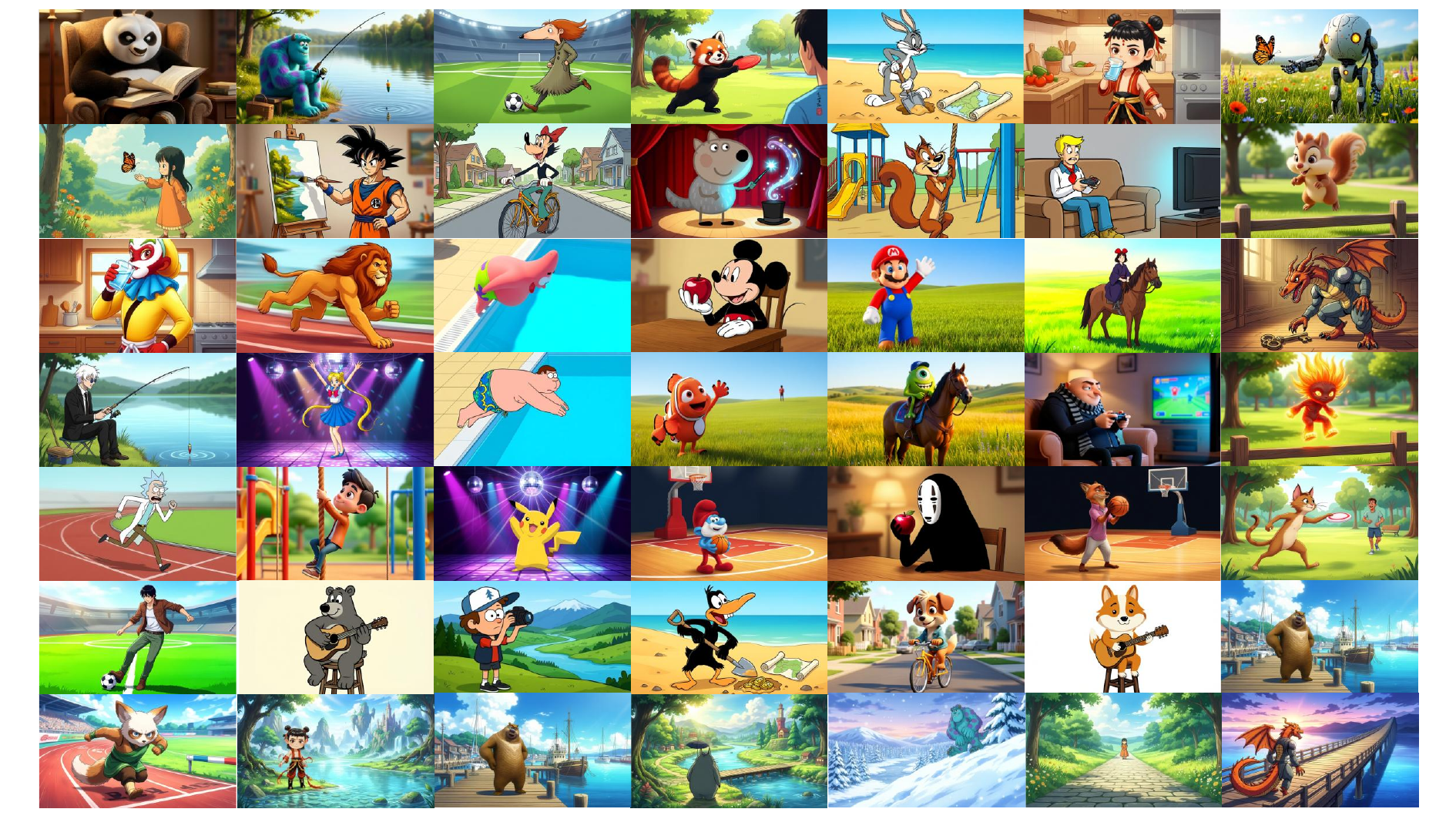}
\caption{\textbf{Examples from our Image Suite.} The figure showcases characters from our proprietary cartoon IP dataset, illustrating a variety of styles and categories.}
\label{fig:image suite}
\end{figure}

\paragraph{Image Suite.} We constructed a proprietary cartoon IP dataset containing detailed character profiles, reference images, and multi-view character sheets. To ensure alignment between visual assets and textual prompts, Qwen-Image-Edit~\cite{wu2025qwen} was employed to modify the background and initial poses of characters, effectively synthesizing the starting frames. The dataset includes 10 existing IPs. However, many of these are protected by copyright, which often prevents closed-source models from generating them. Furthermore, since many models might have encountered these IPs during training, using them for evaluation may not accurately reflect a model's generalization ability. To address these issues, we added 30 self-designed IPs, resulting in a total of 30 2D and 10 3D characters. These assets cover a wide range of categories, such as animals, robots, and humans. They also include various animation styles, including Disney, Japanese Anime, Minimalism, American Comics, and 1990s Retro. Examples of these assets can be seen in Fig.~\ref{fig:image suite}.

\paragraph{Prompt Suite.} For each character, we designed unique descriptions based on classic animation standards, including Appearance Features, Behavioral Logic, and Personality. We also defined specific evaluation principles and requirements for different dimensions, such as scenes, actions, and objects. Then, we combined these two types of prompts and provided them to a VLM along with images sampled from our image suite. The VLM generated a prompt proposal, which was checked by humans and used as the final prompt for video generation.

\begin{figure}[t!]
    \centering
    \includegraphics[width=\linewidth]{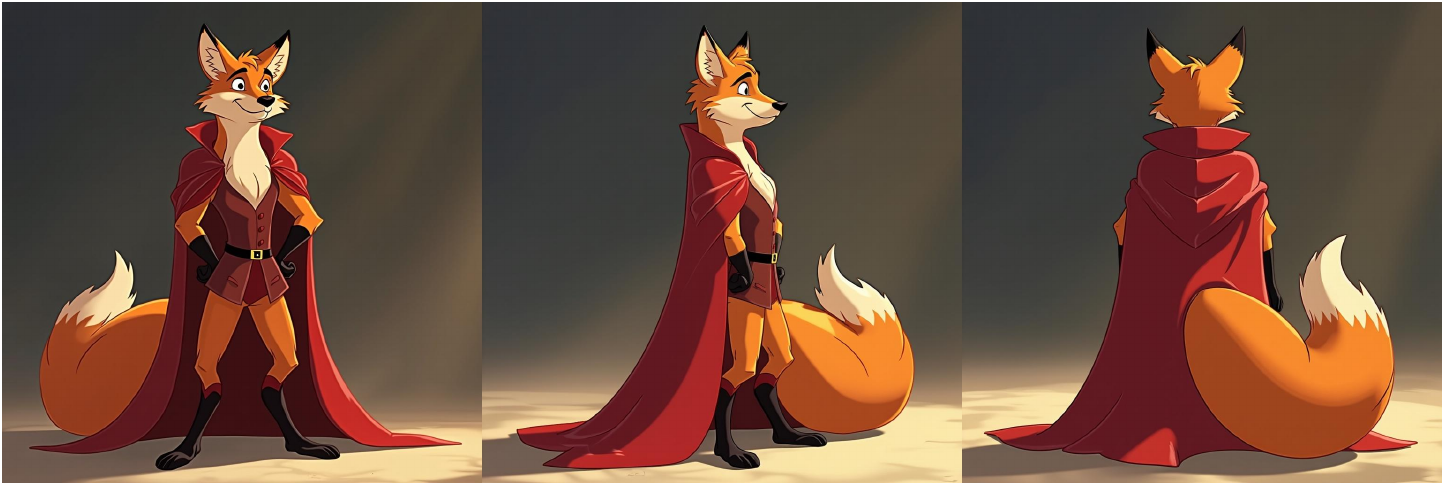}
    \caption{\textbf{Multi-view IP images of Rex the Red Fox} }
    \label{fig:Fox}
    \vspace{-3mm}
\end{figure}

\paragraph{IP Setting.}
In the development of our proprietary Intellectual Property (IP), we synthesized established industry standards from character animation with original creative elements. The character's design framework is structured across three primary dimensions: Appearance, Behavior, and Personality.

Appearance setting comprises general physiological features and orthographic projections (front, side, and back views), supplemented by high-fidelity reference imagery of the character in various environmental contexts. Behavior setting is categorized by signature movements, action logic, and environmental interaction patterns. Personality setting is defined through qualitative trait descriptions and a dedicated 'facial acting' schema to ensure emotional consistency.

A detailed case A detailed case study is presented in Fig.~\ref{fig:Fox} and Table~\ref{tab:rex_character_profile}

\begin{table*}[t]
\centering
\small
\setlength{\tabcolsep}{6pt}
\begin{tabular}{lp{12cm}}
\toprule
\textbf{Attribute} & 
\textbf{Description} \\
\midrule

Character Name & Rex the Red Fox \\
\midrule

\multicolumn{2}{l}{\textbf{Canonical Appearance}} \\
\midrule

Summary & 
A confident anthropomorphic red fox with a vibrant orange coat and a large, bushy tail, dressed in a dramatic red cloak and a matching vest over orange trousers, exuding a heroic and theatrical presence. \\

Front View & 
The character stands in a heroic, confident pose, facing forward with hands on hips. He has a vibrant orange coat with a cream-colored chest and muzzle. His face features large, expressive eyes with long black lashes, a small black nose, and a wide, charming smile. His ears are pointed and tufted with black tips. He wears a long, flowing red cloak with a high collar, a fitted red vest with three visible buttons, and bright orange trousers. His arms are covered in black gloves that extend to his elbows, and he wears black boots. His large, fluffy tail is curled behind him, with a white tip. \\

Side View & 
The character is shown in a three-quarter profile, looking towards the right. His posture is upright and poised. The side view highlights the sleek lines of his body, the voluminous red cloak draped over his shoulder, and the prominent, fluffy tail with its white tip. The orange of his trousers and the red of his vest are clearly visible, along with the black gloves and boots. His head is turned slightly, showing his profile with a confident smile. \\

Back View & 
The character is viewed from behind, standing with his back to the viewer. The high collar of his red cloak is prominent, and the cloak flows down his back and to the sides. His large, bushy tail is curled around his legs, with the white tip clearly visible. The back of his red vest and the top of his orange trousers are seen. His pointed ears are visible at the top of his head, and the overall silhouette is that of a confident, noble figure. \\
\midrule

\multicolumn{2}{l}{\textbf{Canonical Behavior}} \\
\midrule

Personality & 
Confident, charismatic, and theatrical. He carries himself with a regal and self-assured presence, exuding charm and leadership. \\

Action Logic & 
He solves problems with a combination of quick, graceful movements and clever, strategic thinking. His actions are deliberate and often involve using his environment or his cloak for dramatic effect. \\

Signature & 
A slow, deliberate gait with a slight, elegant sway. His movements are fluid and precise, like a seasoned performer. He often uses his cloak as a prop, letting it billow dramatically during key moments. \\

Facial Acting & 
His eyes are highly expressive, capable of conveying a wide range of emotions from playful confidence to focused determination. His smiles are wide and genuine, and his reactions are immediate and clear. \\

Environmental Interaction & 
He interacts with his surroundings with a sense of ownership and control. He might brush a hand across a surface to feel its texture or use his cloak to shield himself or create a moment of distraction. \\

\bottomrule
\end{tabular}
\caption{Character profile for Rex the Red Fox, including canonical appearance details from multiple viewpoints and behavioral characteristics.}
\label{tab:rex_character_profile}
\end{table*}

\begin{table*}[t]
\centering
\caption{\textbf{Number of Evaluated Videos per Model per Dimension in AnimationBench.} Since some videos can be reused across different categories, the total number of unique videos required for generation is 360. This total is composed of the following distributions: Distinctiveness (30), SISO (30), Novelty (30), Semantic (30), Diversity (50), Camera Motion (70), IP (30), S\&S (30), and Physical Laws (including 30 for Anticipation and 30 for Inertia).}
\label{tab:number}
{\setlength{\tabcolsep}{2.2pt}
\renewcommand{\arraystretch}{1.05}
\scriptsize
\resizebox{\textwidth}{!}{
\begin{tabular}{cccccccccccccccccccc}
\toprule

\textbf{App.} & \textbf{Beh.} & \textbf{Pers.}
& \textbf{Antic.} & \textbf{FTOA} & \textbf{SI/SO}
& \textbf{S\&S}
& \textbf{DC} & \textbf{SD} & \textbf{Nov.}
& \textbf{Div.} & \textbf{DD} & \textbf{SE}
& \textbf{Obj.} & \textbf{Act.} & \textbf{Color} & \textbf{Scene}
& \textbf{MR} & \textbf{CMC} \\

\midrule

30 & 30 & 30
& 30 & 30 & 30
& 30
& 30 & 60 & 30
& 50 & 60 & 30
& 30 & 30 & 30 & 30
& 30 & 70 \\

\bottomrule
\label{tab:number_of_generate}
\end{tabular}
}}
\vspace{-3mm}
\end{table*}
\paragraph{Details of Annotator.}
\label{appendix:annotator-details}

Our human evaluation involved 20 expert annotators with professional experience in video generation and animation. For each comparison, annotators were presented with paired model outputs and asked to select the better one for a given evaluation dimension, with an additional tie option when the two results were of comparable quality. To ensure the reliability of the collected judgments, we measured inter-annotator agreement and obtained \textit{Cohen’s Kappa values} above 0.75, indicating strong consistency. In cases of disagreement, the final decision was resolved by a senior animator.
\section{Limitations and Future Work}
One limitation of the current benchmark is that it focuses on visual, character-centric animation quality and does not explicitly assess audio-visual synchrony. This limits its applicability to emerging multimodal generation settings, where alignment between speech, sound, and visual motion is increasingly important. As frontier models continue to advance toward joint audio-visual generation, audio-visual synchrony becomes an important dimension of generation quality. In future work, we plan to extend the benchmark by introducing audio-visual synchrony as an additional evaluation dimension, thereby enabling a more comprehensive assessment of generated character animations.




\end{document}